\documentclass[final]{cvpr}

\usepackage{times}
\usepackage{epsfig}
\usepackage{graphicx}
\usepackage{amsmath}
\usepackage{amssymb}

\usepackage{bigstrut}
\usepackage{subcaption}
\usepackage{bm}
\usepackage{algorithm,lipsum}
\usepackage{wrapfig}
\usepackage[noend]{algpseudocode}
\usepackage[pagebackref=true,breaklinks=true,colorlinks,bookmarks=false]{hyperref}

\newcommand{\myparagraph}[1]{\vspace{6pt}\noindent{\bf #1}}

\def\cvprPaperID{***} % *** Enter the CVPR Paper ID here
\def\confYear{***}

\begin{document}
	
	%%%%%%%%% TITLE
	\title{Generalized Continual Zero-Shot Learning}
	
		\author{Chandan Gautam\\
		Indian Institute of Science, Bangalore\\
		%	Institution1 address\\
		{\tt\small  chandang@iisc.ac.in}
		\and
		Sethupathy Parameswaran\\
		Indian Institute of Science, Bangalore\\
		{\tt\small sethupathyp@iisc.ac.in}
		\and
		Ashish Mishra\\
		IIT Madras\\
		{\tt\small mishra@cse.iitm.ac.in}
		\and
		Suresh Sundaram\\
		Indian Institute of Science, Bangalore\\
		{\tt\small vssuresh@iisc.ac.in}
	}
	
	\maketitle

	%%%%%%%%% ABSTRACT
	\begin{abstract}
		Recently, zero-shot learning (ZSL) emerged as an exciting topic and attracted a lot of attention. ZSL aims to classify unseen classes by transferring the knowledge from seen classes to unseen classes based on the class description. Despite showing promising performance, ZSL approaches assume that the training samples from all seen classes are available during the training, which is practically not feasible. To address this issue, we propose a more generalized and practical setup for ZSL, i.e., continual ZSL (CZSL), where classes arrive sequentially in the form of a task and it actively learns from the changing environment by leveraging the past experience. Further, to enhance the reliability, we develop CZSL for a single head continual learning setting where task identity is revealed during the training process but not during the testing. To avoid catastrophic forgetting and intransigence, we use knowledge distillation and storing and replay the few samples from previous tasks using a small episodic memory. We develop baselines and evaluate generalized CZSL on five ZSL benchmark datasets for two different settings of continual learning: with and without class incremental. Moreover, CZSL is developed for two types of variational autoencoders, which generates two types of features for classification: (i) generated features at output space and (ii) generated discriminative features at the latent space. The experimental results clearly indicate the single head CZSL is more generalizable and suitable for practical applications.
	\end{abstract}
	%\vspace{-10}
	%%%%%%%%% BODY TEXT
	
	%%%%%%%%% BODY TEXT
	\section{Introduction}
	Conventional supervised machine learning (ML)  and, more recently, deep learning algorithms have shown remarkable performance on various tasks (e.g., classification/recognition) in multiple domains, such as Computer Vision and Natural Language Processing. Despite the recent success of supervised ML/deep learning algorithms, they have two crucial limitations: 1- Conventional machine learning models have restricted themselves to training classes. When any example from the novel/unseen class occurs in the test set, such models can not be correctly classified. 2- Conventional machine learning models cannot continually learn over time by accommodating new knowledge from streaming data while retaining previously learned information.
	
	\noindent
	In recent years, the first limitation is addressed by zero-shot learning (ZSL) \cite{aPY,AWA,zeroshotlearning_dataset}, where we classified objects from classes that are not available during training. The second limitation is resolved by continual/lifelong learning \cite{kirkpatrick2017overcoming,riemer2018learning,hayes2019memory}. ZSL approaches having a problem for sequential training, and continual approaches can not handle unseen class objects. Therefore, a more preferable and desirable approach needs to tackle sequential training and unseen object problems simultaneously. This paper aims to leverage the advantages of both zero-shot learning and continual learning in a single framework. Consequently, we propose a single model that can predict the unseen class object and adapt to a new task without forgetting the knowledge about the previous task.
	
	A traditional ZSL method recognizes unseen classes using seen class samples and  textual descriptions of the classes. Despite showing a promising performance of traditional ZSL, it is unable to learn from streaming data. Generative approaches have received surge of interest over embedding based approaches due to the synthetic feature generation ability of unseen class features and a significant improvement in the model's performance. \cite{verma2017simple,wang2018zero,ma2020variational,verma2020meta,kumar2018generalized, xian2019f,huang2019generative,sariyildiz2019gradient,akata2015evaluation, schonfeld2019generalized,xian2019f,verma2020meta}. Therefore, we choose two VAE-based \cite{kingma2013auto} generative methods as baseline for CZSL, namely conditional VAE (CVAE) \cite{mishra2018generative} and cross and distribution aligned VAE (CADA) \cite{schonfeld2019generalized}. CVAE performs ZSL by generating visual features, and CADA performs ZSL by latent feature generation. To enable these two variants of VAEs for continual learning in zero-shot learning framework, we employ memory replay \cite{chaudhry2018efficient,hayes2019memory} with knowledge distillation \cite{hinton2015distilling}. Memory replay is possible in two ways: (a) by replying to real samples of previous tasks \cite{hayes2019memory,chaudhry2019tiny}, and (b) by generating synthetic features for previous tasks \cite{shin2017continual,liu2020generative}. We empirically choose experience replay-based methods (by real samples) over generative feature replay-based methods (by synthetic samples) as generative feature replay-based methods perform inferior compared to experience replay. Moreover, continual learning experiments can be done for two types of settings: single and multi-head settings \cite{chaudhry2018riemannian,liu2020generative}. Single-head setting performs a task-agnostic prediction (i.e., task identifier is unknown at the testing time); however, multi-head settings require samples' task identity during testing. Therefore, a single-head setting is more challenging and plausible for real-world problems. We present all the experiments and analyses in a single-head setting for zero-shot learning in a continual learning framework in this work. The main contributions of this work are summarized as follows:
	
	\begin{enumerate}
		\item Experience replay-based CZSL method is developed using generative models with small episodic memory.
		\item The proposed method is developed for a single-head setting, which performs task-agnostic prediction, and is more convenient to solve a real-world problem.
		\item  CZSL has experimented on two kinds of continual learning settings. One CZSL setting already exists \cite{skorokhodov2020normalization}. The other is proposed for two purposes: (a) to evaluate class-incremental ability of the CZSL method as the incremental class evaluation is not possible on the existing setting \cite{skorokhodov2020normalization};  (b) to compare the performance of the continual model from the offline ZSL models as the train-test split of the last task of the proposed CZSL setting is identical to standard ZSL split.
		\item  Exhaustive experiments are conducted on 5 ZSL benchmark datasets for a continual learning in zero-shot learning framework. The proposed method outperforms the baselines and the existing CZSL methods. 
	\end{enumerate}

	\section{Related Work}
	\subsection{Zero-shot learning}
	Recently, ZSL has attracted considerable attention due to handling unknown objects during testing. It transfers knowledge from seen classes to unseen classes via class attributes. Earlier proposed approaches for ZSL primarily were discriminative or non-generative (i.e., embedding-based) in nature \cite{akata2016label,ConSE, xian2016latent, LampertNH14, zhang2015zero, socher2013zero, CSSL1, CSSL2, CSSL3, CSSL4, CSSL5}. Non-generative methods learn an embedding from visual space to semantic space or vise Versa via a linear compatibility function \cite{akata2016label,ConSE, xian2016latent, LampertNH14}. These approaches are based on the assumption that the class attributes of seen and unseen classes share many similarities. Embedding-based approaches represent image class as a point hence unable to capture intra-class variability.
	
	The generalized ZSL (GZSL) problem is potentially more practical and challenging where the training and the test classes are not disjoint. As ZSL models train using seen class examples, most of the embedding based approaches show a strong bias towards the seen classes in GZSL. VAE and GAN are the backbones for generative models used to synthesize the examples for several applications. The ability to synthesize the seen/unseen class examples from class attributes using VAE/GAN are the basis of generative models-based ZSL.
	Generative models have recently shown promising results for both ZSL and GZSL setups using synthesized examples for unseen classes to train a supervised classifier \cite{vermageneralized,cycle-UWGAN, f-CLSWGAN, DGAN, GZSL_few, ZSL_back, ZSL_side, f-VAEGAN}.  A particular advantage of the generative models is that they transform a ZSL problem to a typical supervised learning problem using the synthesized examples of unseen classes to train a supervised classifier. 	
	\subsection{Continual Learning}
	
	Continual learning learns from streaming data with two objectives: avoid catastrophic forgetting (preserve experience while learning on new tasks) and avoid intransigence (update new knowledge and transfer the previous knowledge). Overall literature of continual learning can be divided into three parts: regularization-based \cite{kirkpatrick2017overcoming,chaudhry2018riemannian,rebuffi2017icarl}, memory-based \cite{lopez2017gradient,chaudhry2018efficient}, experience replay-based \cite{hayes2019memory,chaudhry2019tiny}. This paper is focused on experience replay for a single-head setting.%Generally, continual learning tests under two types of evaluation settings \cite{chaudhry2018riemannian}: single-head and multi-head.
	
	\subsection{Continual Zero-shot Learning} 
	In a traditional continual learning setting, training and testing data contain the same number of classes for classification. However, in the CZSL setting, training data also contains some unseen classes with their description in textual form, and a classifier should be able to classify these unseen classes during testing. Most recently, CZSL \cite{wei2020lifelong,skorokhodov2020normalization} has drawn increasing
	interest. To the best of our knowledge, only a handful
	number of work is available for this problem. Chaudhry et al. \cite{chaudhry2018efficient} develop an average gradient episodic memory (A-GEM) -based CZSL method for a multi-head setting. A generative model-based CZSL \cite{wei2020lifelong} method is also developed for multi-head setting. Most recently, Skorokhodov and Elhoseiny \cite{skorokhodov2020normalization} develop an A-GEM-based CZSL method for a single-head setting.    
	
	\section{Problem Formulation}
	
	In single-head CZSL problem, a data stream contains feature information, task identity, and class label for seen classes; and class attribute information for seen and unseen classes. The $t^{th}$ task for a data stream consists of training and testing  stream. The training stream is defined as: $\mathcal{D}^t_{tr} = \{(x_i^t, \iota_i^t, y_i^t, a_i^t)_{i=1}^{n_{tr}}\}$. Here, $x_i^t$, $\iota_i^t$, $y_i^t$, and $a_i^t$ denote feature vector, task identity, class label, and class attribute information for $i^{th}$ sample of $t^{th}$ task, respectively. $n_{tr}$ denotes number of training samples in $t^{th}$ task. $\mathcal{D}^t_{tr}$ contains seen classes information. Apart of that, class attribute information of unseen classes, $\mathcal{U_C} = \{(a_i)_{i=1}^{n_{uc}}\}$ are also available, where $n_{uc}$ is number of unseen classes. This information enables the model for prediction of unseen testing samples. Further, the testing stream is defined as: $\mathcal{D}^t_{ts} = \{(x_i^t, y_i^t)_{i=1}^{n_{ts}}\}$. In this stream, $x_i^t$ and $y_i^t$ denote feature vector and class label for $i^{th}$ sample of $t^{th}$ task, respectively. $n_{ts}$ denotes number of test samples in $t^{th}$ task. Here, class label of testing data is only used for evaluation purpose. It is to be noted that testing data stream does not contains any task identity as we propose a single-head based method. The proposed method develops a shared model for all tasks and doesn't utilize any task information for prediction.
	
	\begin{figure*}[h!]
		\centering
		\includegraphics[width=1.0\linewidth]{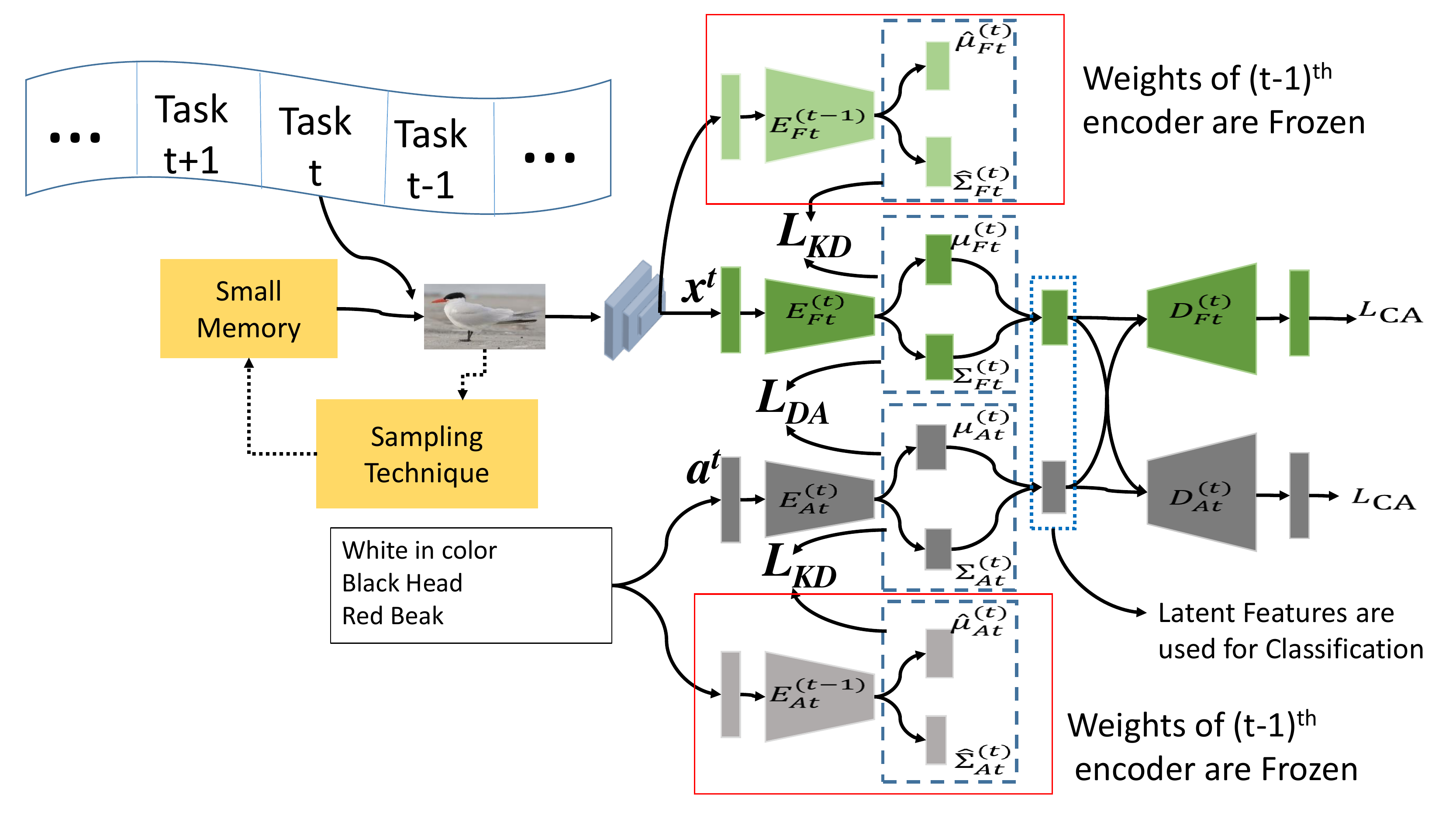}
		\caption{Proposed framework for CZSL-CA using experience replay}
		\label{fig:CZSL-CA}
	\end{figure*}
	
	\section{Generalized Continual Zero-Shot Learning}
	In this section, the continual learning method is proposed for the ZSL framework. As discussed above, CZSL methods are developed for single \cite{skorokhodov2020normalization}, as well as multi-head \cite{chaudhry2018efficient} setting using average gradient episodic memory. These papers utilize the episodic memory for storing the gradient. In our work, we directly train on the samples of the previous tasks, which are stored in small episodic memory. This strategy is called experience replay. Experience replay is a well-established method in the field of continual learning \cite{hayes2019memory,chaudhry2019tiny}. We develop a CZSL method based on experience replay (ER), knowledge distillation, and two variants of variational autoencoder (VAE), namely conditional VAE (CVAE) \cite{mishra2018generative}, and Cross and Distribution Aligned VAE (CADA) \cite{schonfeld2019generalized}.  We describe CADA-based CZSL method (i.e., CZSL-CA) in Section \ref{subsec:CZSL-CA}. Due to space constraints, the CVAE-based CZSL method (i.e., CZSL-CV) is provided in the supplementary material.

	\subsection{Experience Replay in CZSL-CA}\label{subsec:CZSL-CA}
	
	Like other continual learning methods, the primary purpose of CZSL-CA is to preserve previous knowledge and adopt new knowledge at the same time. For this purpose, we employ the ER with CADA. ER utilizes a small episodic memory to replay the data from memory for preserving the previous knowledge and perform a joint training of previous and current samples. Here, the crucial task is to select the samples for storing in the memory. Various memory populating strategies are available in the literature \cite{riemer2018learning,chaudhry2019tiny}. We have utilized three kinds of techniques for this purpose, which are briefly discussed in Section \ref{subsec:sample}, and pseudo-code for these sampling techniques is provided in the supplementary material.
	The complete training framework of CZSL-CA is provided in Figure \ref{fig:CZSL-CA}. A pseudo-code of this training framework is provided in the supplementary material. In Figure \ref{fig:CZSL-CA}. For $t^{th}$ task, two encoders (feature ($E_{Ft}$) and attribute ($E_{At}$) encoders) and two decoders (feature ($D_{Ft}$) and attribute ($D_{At}$) decoders) are trained by minimizing four kind of losses. Three losses are the same as the loss of CADA \cite{schonfeld2019generalized}, and the fourth loss is knowledge distillation loss (KD) between current and previous encoders. All four losses are mentioned below: 
	
	\myparagraph{VAE loss.} It is the sum of two standard VAE loss; one is for feature encoder and decoder, and the other is for attribute encoder and decoder.
	
	\myparagraph{Distribution-alignment loss (DA).} It minimizes the distribution between the latent space information of feature and attribute encoder for each task. 
	\begin{equation}
		\mathcal{L}_{DA} = (\|\mu_{At}^t - \mu_{Ft}^t\|_2^2 + \|(\Sigma_{At}^t)^{\frac{1}{2}} - (\Sigma_{Ft}^t)^{\frac{1}{2}}\|_{\mathcal{F}}^2)^{\frac{1}{2}},
	\end{equation}
	where $\mu_{Ft}^t$ and $\Sigma_{Ft}^t$ are predicted by feature encoder, $\mu_{At}^t$ and $\Sigma_{At}^t$ are predicted by attribute encoder, and $\mathcal{F}$ denotes Frobenius norm. 
	
	\myparagraph{Cross-alignment loss (CA).} It is cross-reconstruction loss between original modality and the modality obtained by decoder for $t^{th}$ task as follows:
	\begin{equation}
		\mathcal{L}_{CA} = |a - D_{At}^t(E_{Ft}^t(x))| + |x - D_{Ft}^t(E_{At}^t(a))|,
	\end{equation}
	where $a$ is the class attribute vector, $D_{At}^t$ is the attribute decoder, $E_{At}^t$ is the attribute encoder, $x$ is the feature vector, $D_{Ft}^t$ is the feature decoder, and $E_{Ft}^t$ is the feature encoder. 
	
	\myparagraph{Knowledge distillation loss (KD).} It minimizes the difference between latent space of $t^{th}$ and $(t-1)^{th}$ feature ($E_{Ft}$), as well as attribute encoder ($E_{At}$). It helps in transferring the previous knowledge. The loss is defined as follows:       
	\begin{equation}
		\mathcal{L}_{KD} = |E_{Ft}^{t}(x) - E_{Ft}^{t-1}(x)|_1 + |E_{At}^{t}(a) - E_{At}^{t-1}(a)|_1,
	\end{equation}
	where $E_{Ft}^{t}$ and $E_{Ft}^{t-1}$ is the feature encoder of the current task  and the previous task, respectively. Similarly $E_{At}^{t}$ and $E_{At}(a)^{t-1}$ is the attribute encoder of the current task and the previous task, respectively.
	
	\noindent
	The overall loss of CZSL-CA is as follows:
	\begin{equation}
		\mathcal{L}_{CZSL-CA} = \mathcal{L}_{VAE} + \gamma \mathcal{L}_{CA} + \delta \mathcal{L}_{DA} + \mathcal{L}_{KD},
	\end{equation}
	where $\gamma$ and $\delta$ are the weighting factors.
	
	During training, a few samples from each task are stored in small memory. Further, except $1^{st}$ task, train data of all tasks are generated by stacking samples from the current task and samples from this small memory. Once all encoders and decoders are trained for the $t^{th}$ task, the trained feature encoder ($E_{Ft}^t$) is used to generate latent features from train data (seen classes) using visual features. Similarly, the trained attribute encoder ($E_{At}^t$) is used to generate latent features for each unseen class of task $t$ using the class attribute information. Finally, we concatenate latent features of seen and unseen classes and train a linear classifier for $t^{th}$ task. It is to be noted that the latent features are generated by sampling based on the mean and variance obtained from the respective encoders. The CZSL-CA framework in Figure \ref{fig:CZSL-CA} shares the same encoders and decoders among all tasks as it is a single-head setting-based method. As these encoders and decoders do not vary if the number of classes varies at each task, they can be easily used for a single-head setting. Only a classifier at the last layer needs to develop for class-incremental learning, which is discussed in Section \ref{subsec:class_inc}. Once the training of CZSL-CA completes, testing is performed at each task. The test data of task $t$ is encoded using the trained feature encoder ($E_{Ft}^t$) and is passed to the trained linear Classifier for prediction. In the whole training procedure, populating memory at each task is crucial, and it is discussed next.

	\subsection{Sampling Techniques}\label{subsec:sample}
	Sampling techniques play a key role in continual learning.
	We use various following sampling techniques with CZSL-CA:
	
	\myparagraph{Reservoir Sampling \cite{riemer2018learning,vitter1985random}.}
	It randomly selects samples from the input data stream. Each sample is selected with a probability $\frac{mem_size}{n}$, where $n$ is the total number of samples presented so far, and $mem_size$ is memory size.
	
	\myparagraph{Ring Buffer \cite{lopez2017gradient}.}
	It uses a separate queue data structure of equal size for each class of a task. Hence, the memory size is equivalent to $C \times queue\_size$, where $C$ is the total number of classes in the dataset. It selects a few samples from the last to populate the memory. In contrast to reservoir sampling, episodic memory consists of equal representation of all classes. 
	
	\myparagraph{Mean of Features \cite{rebuffi2017icarl}.}
	It takes a running average of the features of each class and stores samples whose features are close to the running average. It assigns a separate memory of equal size for each class, like the ring buffer sampling.
	\vspace{-0.4cm}
	\subsection{Class Incremental Learning}\label{subsec:class_inc}
	We are solving a CZSL problem for single-head setting as it provides a task-agnostic prediction. If all data are available for training, then any classifier can learn all of these classes jointly. For the CZSL problem, data comes sequentially in the form of a task. We generate latent features using $t^{th}$ encoder of CZSL-CA for $t^{th}$ task and use these latent features for classification. These latent features are beneficial in two ways: (a) we can generate as many samples we want for underrepresented classes; (b) these latent features are quite discriminative and can be easily classified. Since latent features are discriminative, a simple linear classifier with softmax is used for classification. For a new class at $t^{th}$ task, a trained linear classifier of $(t-1)^{th}$ task can be easily and efficiently extended by extending softmax as mentioned in \cite{liu2020generative}.

	\section{Experiments}\label{sec:exp}
	
	We have evaluated our model in five benchmark datasets used for Zero-Shot Learning,  namely Attribute Pascal and Yahoo (aPY) \cite{aPY}, Animals with Attributes (AWA1 and AWA2) \cite{aPY}, Caltech-UCSD-Birds 200-2011 (CUB) \cite{CUB}, and SUN \cite{SUN}. CZSL settings and their evaluation metrics for these datasets are discussed in the subsequent section.
	
	%For all datasets, we use the 2048-dim top-layer pooling units of the 101-layered ResNet \cite{resnet} for feature extraction as suggested by \cite{zeroshotlearning_dataset}.
	
	% 	\begin{table}[h]
	% 		\centering
	% 		\begin{tabular}{|l|p{1.6cm}|p{1cm}|p{1cm}|p{1cm}|}
	% 		\hline
	% 			Dataset & Attribute Dimension & Seen Classes & Unseen Classes & Total Classes   \\
	% 			\hline
	% 			CUB &  312 &  150 &  50 &  200\\
	% 			aPY & 64 & 20 & 12 & 32\\
	% 			AWA1  & 85 & 40 & 10 & 50\\
	% 			AWA1  & 85 & 40 & 10 & 50\\
	% 			SUN  & 102 & 645 & 72 & 717\\
	% 			\hline
	% 		\end{tabular}
	% 		\caption{Standard Split of ZSL Datasets}
	% 		\label{tab_std_split}
	% 	\end{table}
	
	\subsection{Datasets, Settings and Evaluation Metrics} \label{subsec:exp_aval}
	In this section, we discuss two different settings and their corresponding evaluation metrics, which are used to evaluate the proposed Continual Zero-Shot Learning methods as follows:
	
\myparagraph{CZSL Setting-1:}
	This setting closely follows the setting used in \cite{skorokhodov2020normalization}. It decides whether a class is seen or unseen based on the number of tasks trained so far. If method is training on $t^{th}$ task then all classes till $t^{th}$ task are assumed as seen classes, and remaining classes after $t^{th}$ task are assumed as unseen classes for $t^{th}$ task. This setting is more like a traditional ZSL setting. The following metrics are used to evaluate the performance for continual zero-shot learning after $t^{th}$ task \cite{skorokhodov2020normalization}:
	\begin{itemize}\itemsep-0.5em
		\item Mean Seen Accuracy (mSA)
		\begin{equation}
			mSA = \frac{1}{T}\sum_{t=1}^T CAcc(\mathcal{D}_{t_s}^{\leq t}, A^{\leq t}),
		\end{equation}
		where $CAcc$ stands for per class accuracy. 
		\item Mean Unseen Accuracy(mUA)
		\begin{equation}
			mUA = \frac{1}{T-1}\sum_{t=1}^{T-1} CAcc(\mathcal{D}_{t_s}^{> t}, A^{> t})
		\end{equation}
		\item Mean Harmonic Accuracy (mH)
		\begin{equation}
			mH = \frac{1}{T-1}\sum_{t=1}^{T-1} H(\mathcal{D}_{t_s}^{\leq t}, \mathcal{D}_{ts}^{> t}, A),
		\end{equation}
		where H stands for harmonic mean.
		%		\item Mean Joint Accuracy(mJA)
		%		\begin{equation}
		%		mJA = \frac{1}{T}\sum_{t=1}^{T} Acc(D_{t_s}, A)
		%		\end{equation}
		
	\end{itemize}
	
	Here, $T$ denotes the total number of tasks, $\mathcal{D}_{t_s}^t$ denotes the test data of $t^{th}$ task from seen classes,  $\mathcal{D}_{t_{us}}^t$ denotes the unseen test data of $t^{th}$ task, and $A$ denotes the set of all class attributes. 
	Further, $\mathcal{D}^{\leq t}$ denotes all train/test data from $1^{st}$ to $t^{th}$ task, and
	$D^{> t}$ denotes all train/test data from $(t+1)^{th}$ to last task. We also compute the forgetting measure \cite{lopez2017gradient,chaudhry2018efficient} to evaluate	the capability of the model to acquire new knowledge without catastrophic forgetting.
	
	% 	Dataset division as per setting-1: For 200 classes of the CUB dataset, it splits into ten tasks of 10 classes each. Similarly, the aPY dataset, which contains 32 classes, splits into four tasks with eight classes per task. For the AWA1 and AWA2 datasets, which have 50 classes, each is split into five tasks with ten classes per task. The SUN dataset has 717 classes, and it is difficult to split evenly. Hence, it is split into 15 tasks with 47 classes in the first three tasks and 48 classes in the remaining tasks. For all datasets, 20 percent of data from each task is taken as test data to compute the final evaluation metrics.
	
	\myparagraph{CZSL Setting-2.}
	In the first setting, seen and unseen classes are decided based on the number of tasks that have been trained so far. All classes from the trained task are treated as seen, and classes from the remaining tasks are treated as unseen classes. Since classes from all tasks are involved during prediction at each task, this setting can not be used for the class-incremental setup of continual learning. Setting-1 is closer to the ZSL setup than the continual learning setup, as it is impossible to get the class attribute information from all tasks apriori. We propose another setting for CZSL, which is closer to the continual learning like setup. Each task in this setup contains both seen and unseen classes based on the standard split of ZSL \cite{zeroshotlearning_dataset}. In this setting, the test data consists of the unseen classes and 20 percent data from each task's seen classes. The model can be trained/tested only with the classes of the current and previous tasks. Classes of upcoming tasks are not exposed to the model as they are still unknown. If some classes are seen/unseen at one task in this setting, then it will be seen/unseen throughout all tasks. When we arrive at the last task, we have the same number of seen/unseen classes as the standard split in ZSL. Hence, the performance of the CZSL model at the last task should be equal to the performance of the offline ZSL model on the standard ZSL split. This setting suits both continual learning and ZSL and it can be used for any continual learning setup. The evaluation metrics for setting-2  are as follows:
	
	\begin{itemize} \itemsep-1.5em
		\item Mean Seen Accuracy(mSA)
		\begin{equation}
			mSA = \frac{1}{T}\sum_{t=1}^T CAcc(\mathcal{D}_{t_s}^{\leq t}, A^{\leq t}_{t_s})
		\end{equation}
		\item Mean Unseen Accuracy(mUA)
		\begin{equation}
			mUA = \frac{1}{T}\sum_{t=1}^{T} CAcc(\mathcal{D}_{t_{us}}^{\leq t}, A^{\leq t}_{t_{us}})
		\end{equation}
		\item Mean Harmonic Accuracy(mH)
		\begin{equation}
			mH = \frac{1}{T}\sum_{t=1}^{T} H(\mathcal{D}_{t_s}^{\leq t}, \mathcal{D}_{t_{us}}^{\leq t}, A^{\leq t})
		\end{equation}
		%		\item Mean Joint Accuracy(mJA)
		%		\begin{equation}
		%		mJA = \frac{1}{T}\sum_{t=1}^{T} Acc(D_{t_s}, D_{t_{us}}, A)
		%		\end{equation}	
	\end{itemize}

	Description of all datasets' split for setting-1 and setting-2 is provided in the supplementary material due to space constraint.  	
	% 	Dataset division as per setting-2: Similar to CUB dataset as discussed above, the aPY dataset is split into four tasks, with each task containing five seen classes and three unseen classes. The AWA1 and AWA2 split into five tasks, with eight seen classes and two unseen classes per task. The SUN dataset is split into 15 tasks, with the first three tasks containing 43 seen classes and four unseen classes and the remaining tasks containing 43 seen classes and five unseen classes each.

	%	Where $T$ denotes the total number of tasks, $D_{t_s}^t$ denotes the seen test data of task $t$, $D_{t_{us}}^t$ denotes the unseen test data of task $t$ and $A$ denotes the set of all attributes. Further
	%	$D^{\leq t} = \bigcup_{\tau=1}^t D^{\tau}$, \hspace{0.5cm}
	%	$D^{> t} = \bigcup_{\tau=t+1}^T D^{\tau}$

	\subsection{Baselines}
	
	The research on the intersection between ZSL and CL is less explored. To our best knowledge, there are three methods proposed for continual ZSL. Two works \cite{chaudhry2018efficient,wei2020lifelong} are based on a multi-head setting and, third \cite{skorokhodov2020normalization} is based on a single-head setting. Since a multi-head setting is not feasible for practical scenario \cite{chaudhry2018riemannian}, we focused on a single-head setting. Therefore, we include only a single-head setting-based method \cite{skorokhodov2020normalization} for a fair comparison.  We develop the following baselines to compare the proposed CSZL method.

	\begin{itemize}
		\item \textbf{AGEM+CZSL \cite{skorokhodov2020normalization}:} It is an average gradient episodic memory-based continual ZSL method. The authors have experimented for generalized CZSL setting-1 and provided results only for CUB and SUN datasets.

		\item \textbf{Sequential baseline:} The CVAE and CADA are originally proposed for ZSL.  To develop the baseline for the CZSL framework, we train CVAE and CADA sequentially (Seq-CVAE and Seq-CADA) without considering any continual learning strategy. We initialize weights of the model at $t^{th}$ task by the weights of the model trained at $(t-1)^{th}$ task. After training the Seq-CVAE on the current task, synthetic samples are generated for all the classes which need to classify by using class attribute information. Likewise, for Seq-CADA, latent features are generated for all classes which need to classify. The latent features of seen classes from the current task are generated using trained feature encoder ($E_{Ft}$), and latent features of all remaining classes, which need to classify, are generated using trained attribute encoder ($E_{At}$). These generated features are used to train the classifier. As our proposed methods are developed based on CVAE and CADA, we consider the sequential version of these two methods as baseline methods. 
	\end{itemize}

	\subsection{Results}
\begin{figure*}[h!]
		\centering
		\begin{subfigure}{.5\textwidth}
			\centering
			\includegraphics[width=1.1\linewidth,height=3.6cm]{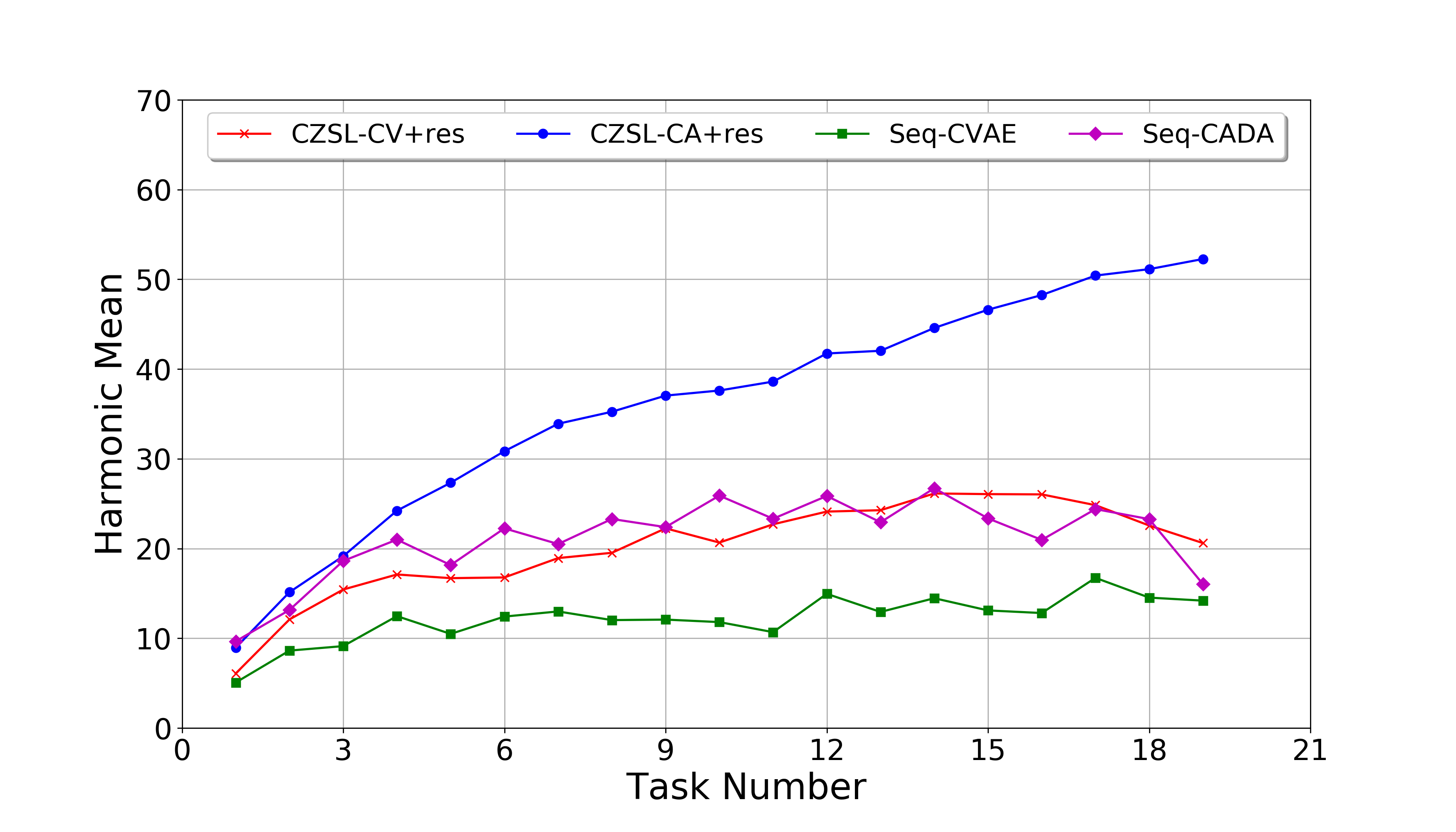}
			\caption{For setting-1}
			\label{fig:cub-mH_s1}
		\end{subfigure}
		\begin{subfigure}{.49\textwidth}
			\centering
			\includegraphics[width=1.1\linewidth,height=3.6cm]{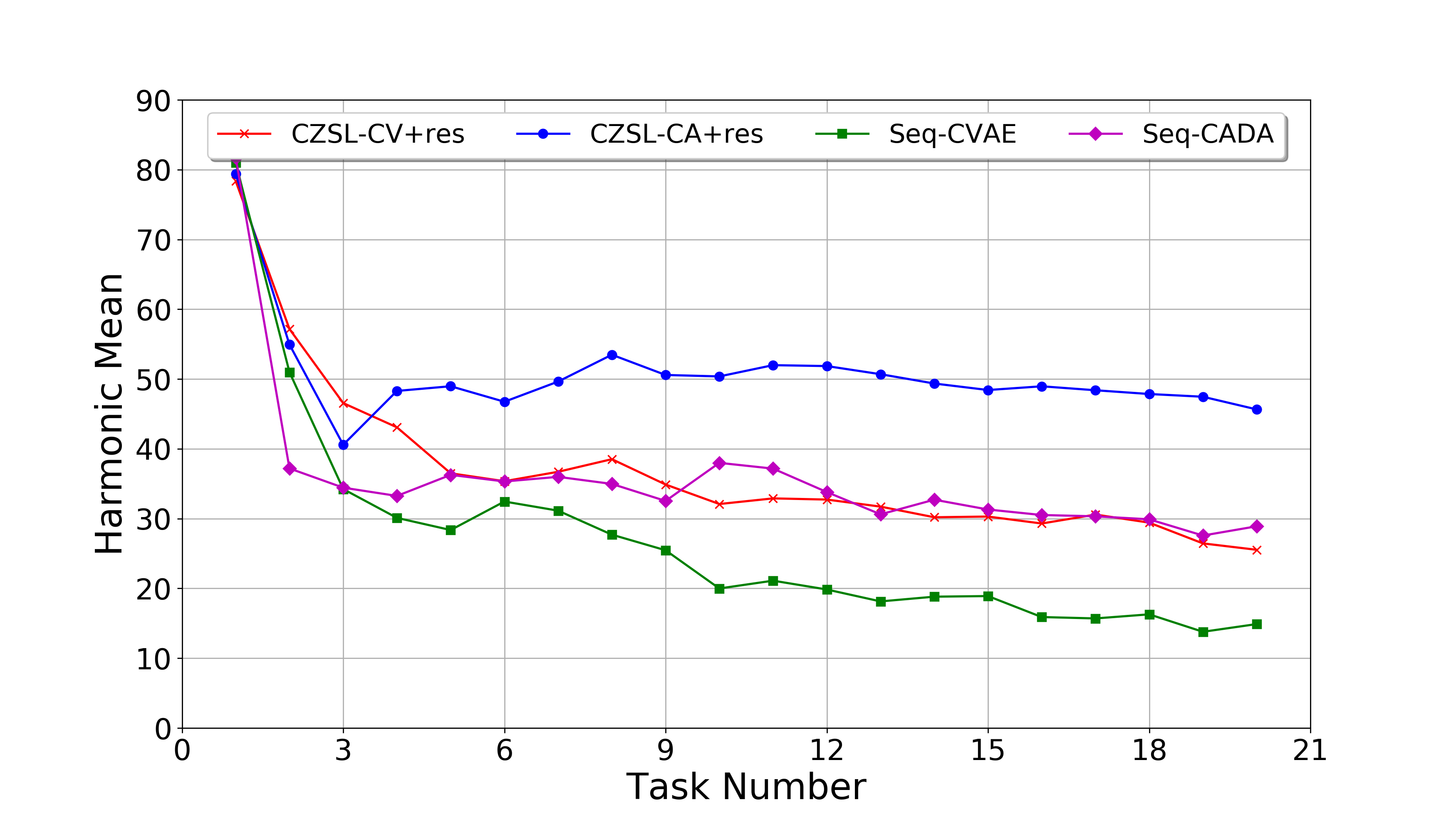}
			\caption{For setting-2}
			\label{fig:cub-mH_s2}
		\end{subfigure}
		\caption{Harmonic mean for CUB dataset after each task for setting-1 and setting-2.}
		\label{fig:taskwise-mH}
	\end{figure*}

	We have performed extensive experiments for Generalized CZSL on two settings of continual learning: CZSL Setting-1 and Setting-2, as mentioned in section \ref{subsec:exp_aval}). The performance  and evaluation of the proposed model in these setups are discussed below:    
	
	\myparagraph{Generalized CZSL for Setting-1.} Results of this evaluation strategy are presented in Table \ref{tab:gen_res_S1} and forgetting measure is provided in supplementary material. The proposed methods significantly outperform baseline methods for all datasets. Among all sampling strategies, The reservoir sampling-based strategy yields either the best results or closer to the best one, and the mean of features-based strategy is the worst performer. It can be analyzed from Figure \ref{fig:taskwise-mH}, performance improves with the task-increment in this setting, because the number of seen classes increases (number of unseen classes decreases) with the tasks-increment, Moreover, CZSL-CV+res exhibits more than $12\%$ improvement over baseline for all datasets except SUN. This improvement is gained by repetitive joint training of samples from the previous task using a minimal episodic memory and the samples from the current task. Here, it is to be noted that repetitive training does not have any adverse effect on the model because $(t+1)^{th}$ task acts as a strong data-dependent regularize for the $t^{th}$ task. 
	
	\myparagraph{Generalized CZSL for Setting-2.} Results of this evaluation strategy are presented in Table \ref{tab:gen_res_S2} and forgetting measure is provided in supplementary material. Similar to generalized CZSL Setting-1, this strategy also yields either the best results or closer to the best one. All performance measures yield better results than setting-1 because the number of unseen classes in setting-1 at initial tasks is more than setting-2. Therefore, low performance on the initial task propagates till the last task in setting-1.

	%also depend upon Setting-1 and Setting-2	
	
	{
		\renewcommand{\arraystretch}{1.4} 
		\begin{table*}[t]
			\centering
			\resizebox{\textwidth}{!}{%
				\begin{tabular}{|l|ccc|ccc|ccc|ccc|ccc|}
					\hline
					& \multicolumn{3}{c|}{CUB} & \multicolumn{3}{c|}{aPY} & \multicolumn{3}{c|}{AWA1} & \multicolumn{3}{c|}{AWA2} & \multicolumn{3}{c|}{SUN} \bigstrut[b]\\
					\cline{2-16}          & \multicolumn{1}{p{4.215em}}{mSA} & \multicolumn{1}{p{2.215em}}{mUA} & \multicolumn{1}{p{2.215em}|}{mH} & \multicolumn{1}{p{2.215em}}{mSA} & \multicolumn{1}{p{2.215em}}{mUA} & \multicolumn{1}{p{2.215em}|}{mH} & \multicolumn{1}{p{2.215em}}{mSA} & \multicolumn{1}{p{2.215em}}{mUA} & \multicolumn{1}{p{2.215em}|}{mH} & \multicolumn{1}{p{2.215em}}{mSA} & \multicolumn{1}{p{2.215em}}{mUA} & \multicolumn{1}{p{2.215em}|}{mH} & \multicolumn{1}{p{2.215em}}{mSA} & \multicolumn{1}{p{2.215em}}{mUA} & \multicolumn{1}{p{2.215em}|}{mH} \bigstrut\\
					\hline
					Seq-CVAE & 24.66 & 8.57  & 12.18 & 51.57 & 11.38 & 18.33 & 59.27 & 18.24 & 27.14 & 61.42 & 19.34 & 28.67 & 16.88 & 11.40 & 13.38 \bigstrut[t]\\
					Seq-CADA & 40.82 & 14.37 & 21.14 & 45.25 & 10.59 & 16.42 & 51.57 & 18.02 & 27.59 & 52.30 & 20.30 & 30.38 & 25.94 & 16.22 & 20.10 \\
					AGEM+CZSL & --  & --  & 13.20 & --  & --  & --  & --  & --  & --  & --  & --  & --  & --  & --  & 10.50 \bigstrut[b]\\
					\hline
					CZSL-CV+res & \textbf{44.89} & 13.45 & 20.15 & 64.88 & 15.24 & 23.90 & \textbf{78.56} & 23.65 & 35.51 & \textbf{80.97} & 25.75 & 38.34 & 23.99 & 14.10 & 17.63 \bigstrut[t]\\
					CZSL-CA+res & 43.96 & \textbf{32.77} & \textbf{36.06} & 57.69 & \textbf{20.83} & \textbf{28.84} & 62.64 & \textbf{38.41} & 45.38 & 62.80 & 39.23 & 46.22 & 27.11 & \textbf{21.72} & \textbf{22.92} \\
					CZSL-CV+rb & 42.97 & 13.07 & 19.53 & 64.45 & 11.00 & 18.60 & 77.85 & 21.90 & 33.64 & 80.92 & 24.82 & 37.32 & 22.59 & 13.74 & 16.94 \\
					CZSL-CA+rb & 43.76 & 31.38 & 35.23 & 61.62 & 19.39 & 27.67 & 65.52 & 38.34 & \textbf{46.45} & 67.09 & 39.41 & \textbf{47.69} & 25.86 & 21.30 & 22.36 \\
					CZSL-CV+mof & 43.73 & 10.26 & 16.34 & \textbf{64.91} & 10.79 & 18.27 & 76.77 & 19.26 & 30.46 & 79.11 & 24.41 & 36.60 & 23.21 & 13.20 & 16.71 \\
					CZSL-CA+mof & 43.95 & 18.79 & 25.36 & 56.62 & 17.80 & 25.84 & 57.31 & 35.10 & 41.59 & 59.13 & \textbf{40.89} & 45.77 & \textbf{27.32} & 16.63 & 20.12 \bigstrut[b]\\
					\hline
				\end{tabular}%
			}
			\vspace{1mm}
			\caption{We provide mean seen accuracy (mSA) for seen, mean unseen accuracy (mUA) for unseen classes, and their mean of harmonic mean (mH) for generalized CZSL setting-1. The best results in the table are presented in bold face.}
			\label{tab:gen_res_S1}%
		\end{table*}%
	}

	{
		\renewcommand{\arraystretch}{1.4} 
		\begin{table*}[t]
			\centering
			\resizebox{\textwidth}{!}{%
				\begin{tabular}{|l|ccc|ccc|ccc|ccc|ccc|}
					\hline
					& \multicolumn{3}{c|}{CUB} & \multicolumn{3}{c|}{aPY} & \multicolumn{3}{c|}{AWA1} & \multicolumn{3}{c|}{AWA2} & \multicolumn{3}{c|}{SUN} \bigstrut[b]\\
					\cline{2-16}          & \multicolumn{1}{p{4.215em}}{mSA} & \multicolumn{1}{p{2.215em}}{mUA} & \multicolumn{1}{p{2.215em}|}{mH} & \multicolumn{1}{p{2.215em}}{mSA} & \multicolumn{1}{p{2.215em}}{mUA} & \multicolumn{1}{p{2.215em}|}{mH} & \multicolumn{1}{p{2.215em}}{mSA} & \multicolumn{1}{p{2.215em}}{mUA} & \multicolumn{1}{p{2.215em}|}{mH} & \multicolumn{1}{p{2.215em}}{mSA} & \multicolumn{1}{p{2.215em}}{mUA} & \multicolumn{1}{p{2.215em}|}{mH} & \multicolumn{1}{p{2.215em}}{mSA} & \multicolumn{1}{p{2.215em}}{mUA} & \multicolumn{1}{p{2.215em}|}{mH} \bigstrut\\
					\hline
					Seq-CVAE & 38.95 & 20.89 & 26.74 & 65.87 & 17.90 & 25.84 & 70.24 & 28.36 & 39.32 & 73.71 & 26.22 & 36.30 & 29.06 & 21.33 & 24.33 \bigstrut[t]\\
					Seq-CADA & 55.55 & 26.96 & 35.62 & 61.17 & 21.13 & 26.37 & 78.12 & 35.93 & 47.06 & 79.89 & 36.64 & 47.99 & 42.21 & 23.47 & 29.60 \bigstrut[b]\\
					\hline
					CZSL-CV+res & 63.16 & 27.50 & 37.84 & \textbf{78.15} & 28.10 & 40.21 & 85.01 & 37.49 & 51.60 & 88.36 & 33.24 & 47.89 & 37.50 & 24.01 & 29.15 \bigstrut[t]\\
					CZSL-CA+res & \textbf{68.18} & \textbf{42.44} & \textbf{50.68} & 66.30 & \textbf{36.59} & 45.08 & 81.86 & 61.39 & \textbf{69.92} & 82.19 & 55.98 & 65.95 & \textbf{47.18} & 30.30 & \textbf{34.88} \\
					CZSL-CV+rb & 60.20 & 27.46 & 37.17 & 77.12 & 25.53 & 36.83 & \textbf{85.93} & 36.09 & 50.32 & \textbf{88.38} & 36.44 & 51.10 & 36.24 & 23.96 & 28.59 \\
					CZSL-CA+rb & 66.01 & 41.72 & 49.95 & 64.88 & 36.04 & \textbf{45.69} & 79.15 & \textbf{62.47} & 69.40 & 83.27 & \textbf{59.80} & \textbf{69.34} & 43.45 & \textbf{31.13} & 34.54 \\
					CZSL-CV+mof & 58.85 & 24.91 & 34.34 & 70.77 & 23.37 & 33.10 & 83.15 & 33.40 & 46.87 & 86.75 & 30.03 & 44.00 & 35.78 & 21.89 & 26.99 \\
					CZSL-CA+mof & 63.83 & 24.10 & 34.22 & 58.43 & 31.32 & 39.72 & 73.05 & 59.84 & 64.42 & 77.72 & 57.28 & 65.37 & 43.86 & 24.06 & 30.75 \bigstrut[b]\\
					\hline
				\end{tabular}%
			}	
			\vspace{1mm}
			\caption{We provide mean seen accuracy (mSA) for seen, mean unseen accuracy (mUA) for unseen classes, and their mean of harmonic mean (mH) for generalized CZSL setting-2. The best results in the table are presented in bold face.}
			\label{tab:gen_res_S2}%
		\end{table*}%
	}	
	
	{
		\renewcommand{\arraystretch}{1.4} 
		\begin{table*}[h!]
			\centering
			\resizebox{\textwidth}{!}{%
				\centering
				
				\begin{tabular}{|l|cc|cc|cc|cc|cc|}
					\hline
					& \multicolumn{2}{c|}{CUB} & \multicolumn{2}{c|}{aPY} & \multicolumn{2}{c|}{AWA1} & \multicolumn{2}{c|}{AWA2} & \multicolumn{2}{c|}{SUN} \bigstrut[b]\\
					\cline{2-11}
					& CZSL-CV & CZSL-CA & CZSL-CV & CZSL-CA & CZSL-CV & CZSL-CA & CZSL-CV & CZSL-CA & CZSL-CV & CZSL-CA \bigstrut\\
					\hline
					Offline (Upper bound) & 34.50 & 52.40 & 22.38 & 40.10 & 47.20 & 64.10 & 51.20 & 63.90 & 26.7  & 40.6 \bigstrut[t]\\
					Sequential (Lower bound) & 14.89 & 28.90 & 0.32  & 6.60  & 34.97 & 23.86 & 26.38 & 26.56 & 16.85 & 23.44 \\
					Reservoir sampling (res) & 25.40 & 45.65 & 23.16 & 29.24 & 39.34 & 60.80 & 38.16 & 62.81 & 21.25 & 32.21 \\
					Ring buffer (rb) & 26.33 & 46.19 & 17.00 & 33.64 & 37.14 & 63.73 & 44.75 & 63.77 & 20.43 & 32.35 \\
					Mean of features (mof) & 22.20 & 27.84 & 11.44 & 31.78 & 31.72 & 58.09 & 34.12 & 58.69 & 19.64 & 27.73 \bigstrut[b]\\
					\hline
				\end{tabular}%
			}	
			\vspace{1mm}
			\caption{Performance comparison of the CZSL methods at the last task and their corresponding offline model (i.e., when we present all tasks training and testing data at once)}
			\label{tab:batch_vs_online}%
		\end{table*}%
	}

	%\myparagraph{Increasing Number of Latent Features.}
	
	% \begin{figure*}
	% 	\centering
	% 	\begin{subfigure}{.5\textwidth}
	% 		\centering
	% 		\includegraphics[width=1.1\linewidth]{./Results/Generalized/Setting-1/All_Methods_UnSeen.png}
	% 		\caption{For setting-1}
	% 		\label{fig:cub-mUA_s1}
	% 	\end{subfigure}
	% 	\begin{subfigure}{.49\textwidth}
	% 		\centering
	% 		\includegraphics[width=1.1\linewidth]{./Results/Generalized/Setting-2/All_Methods_UnSeen.png}
	% 		\caption{For setting-2}
	% 		\label{fig:cub-mUA_s2}
	% 	\end{subfigure}
	% 	\caption{Per class unseen accuracy for CUB dataset after each task for setting-1 and setting-2.}
	% 	\label{fig:taskwise-mUA}
	% \end{figure*}

	\begin{figure*}[h!]
		\centering
		\begin{subfigure}{.5\textwidth}
			\centering
			\includegraphics[width=1.1\linewidth,height=3.8cm]{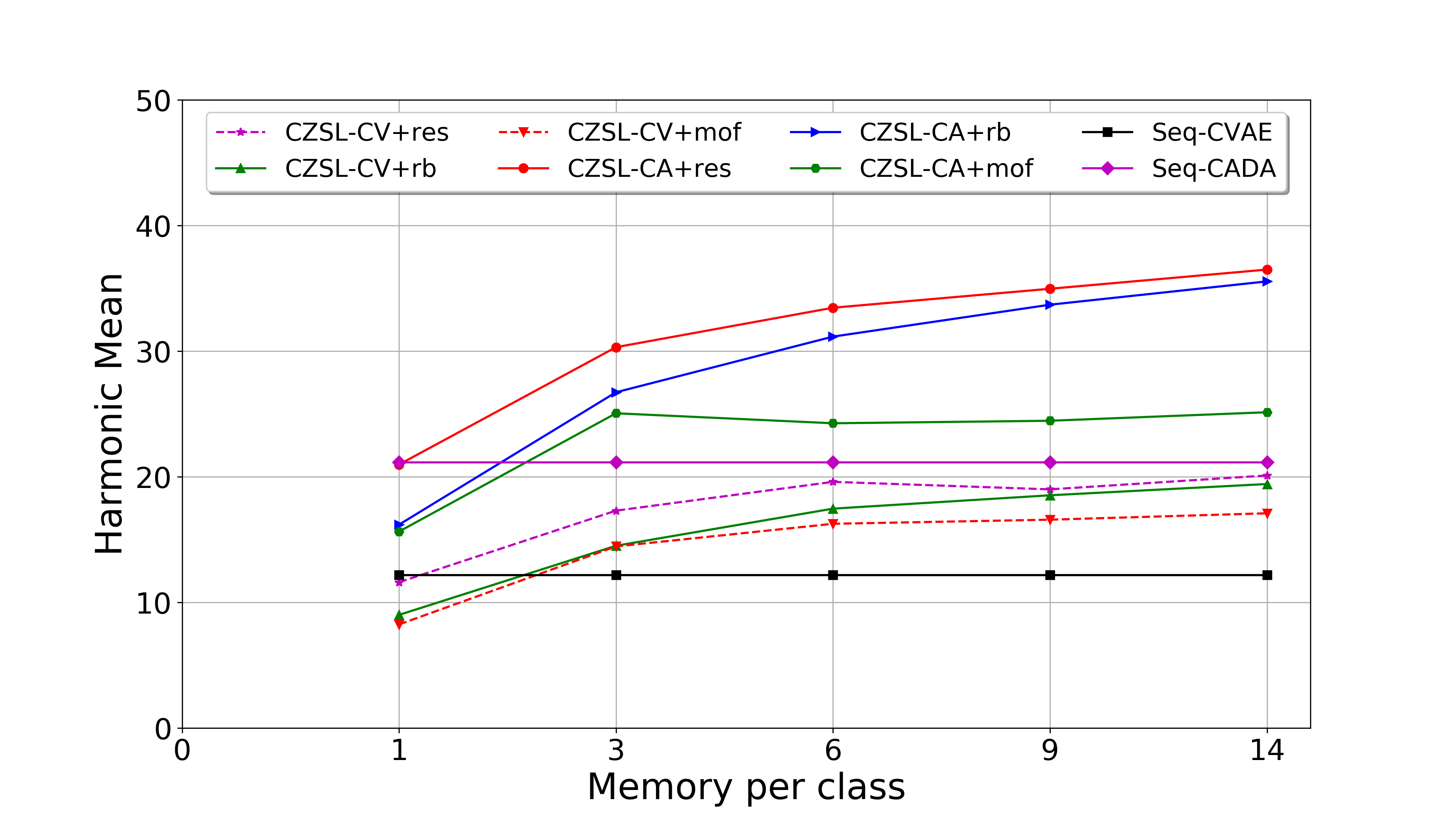}
			\caption{For setting-1}
			\label{fig:abal-mem_s1}
		\end{subfigure}
		\begin{subfigure}{.49\textwidth}
			\centering
			\includegraphics[width=1.1\linewidth,height=3.8cm]{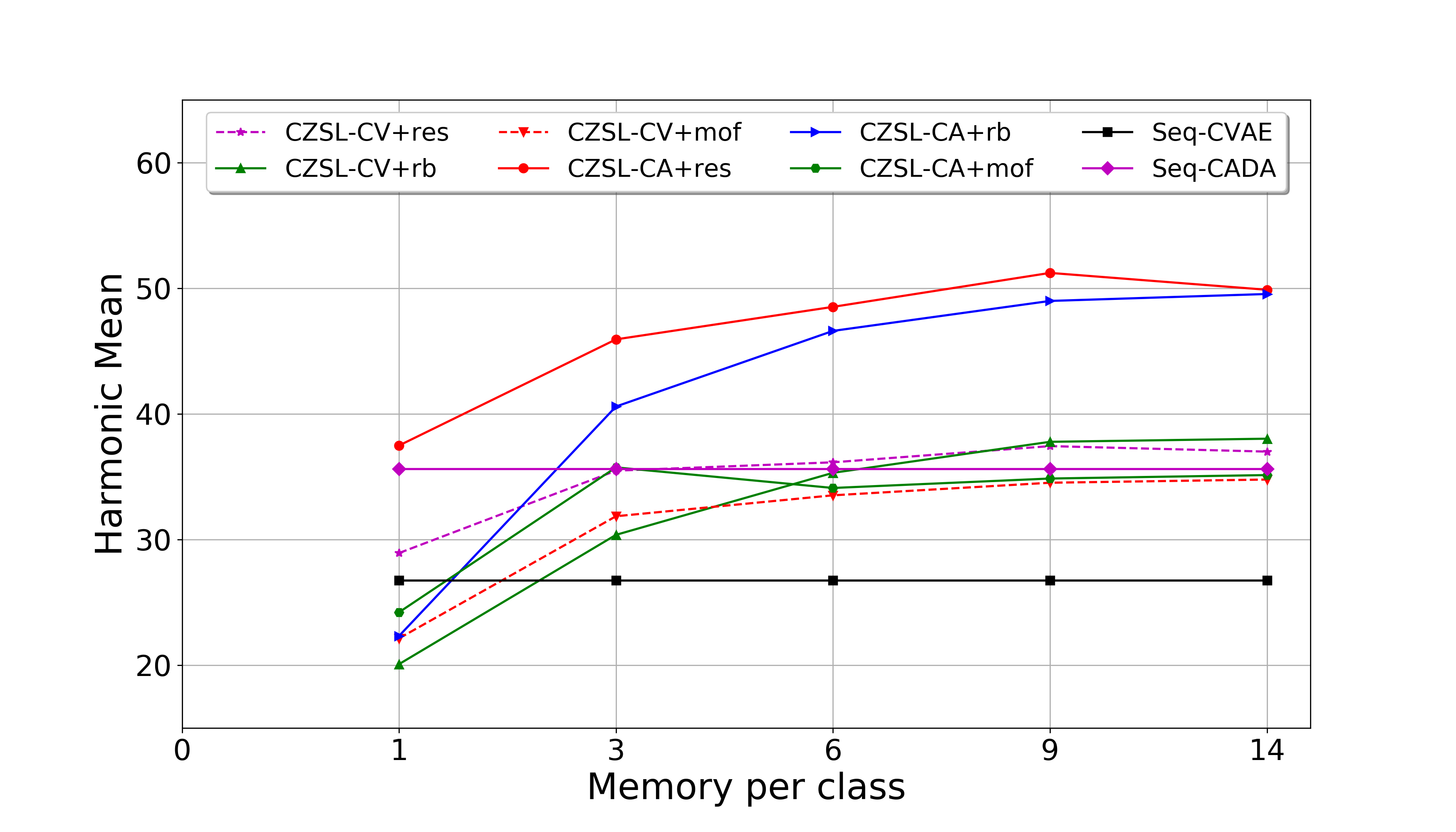}
			\caption{For setting-2}
			\label{fig:abal-mem_s2}
		\end{subfigure}
		\caption{The impact of the number of samples per class in the memory on CUB dataset, which are stored during training of previous task and utilized for training the next task.}
		\label{fig:abal-mem}
	\end{figure*}
	
	\begin{figure*}[h!]
		\centering
		\begin{subfigure}{.5\textwidth}
			\centering
			\includegraphics[width=1.1\linewidth,height=3.8cm]{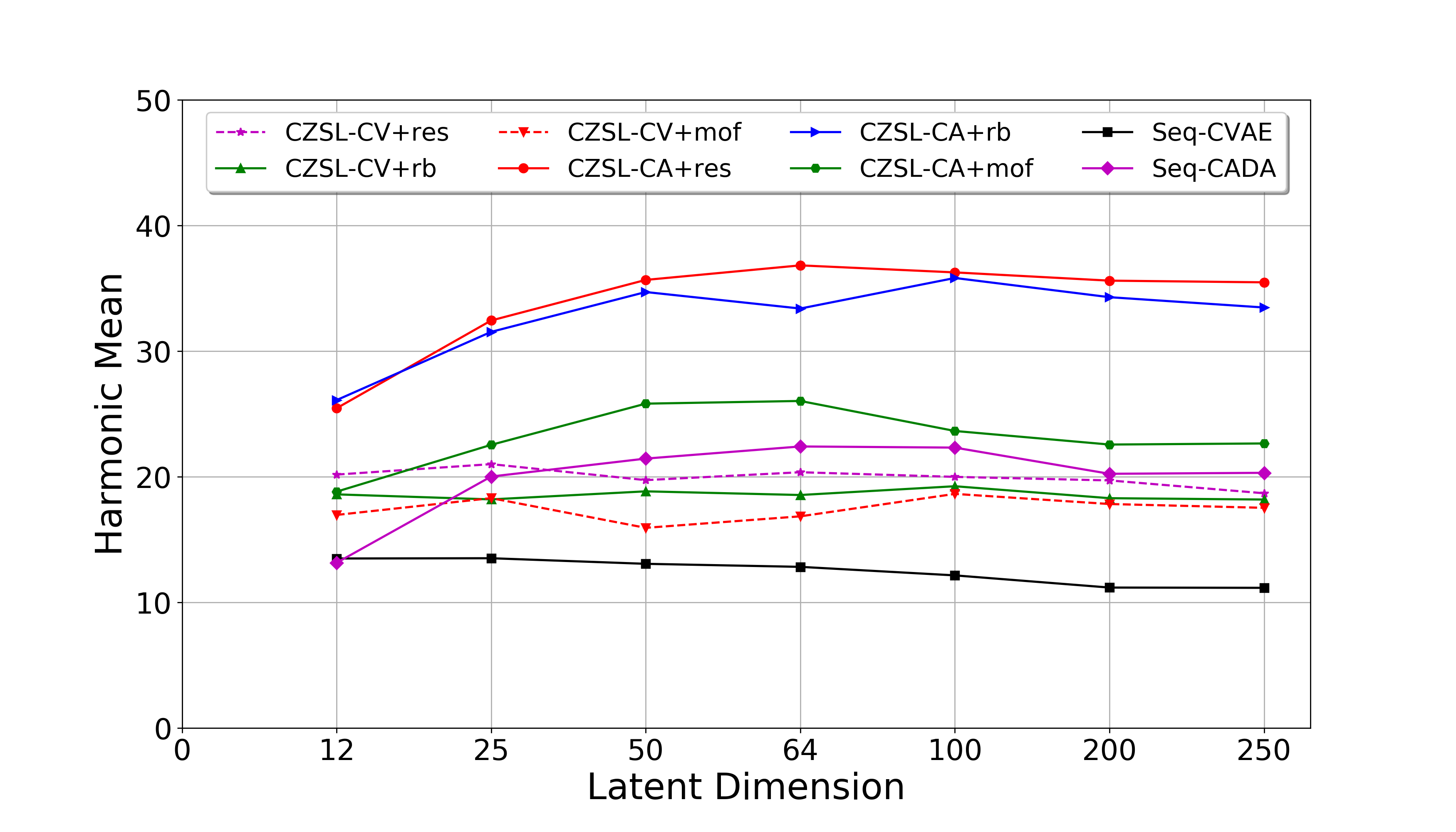}
			\caption{For setting-1}
			\label{fig:abal-lat_s1}
		\end{subfigure}
		\begin{subfigure}{.49\textwidth}
			\centering
			\includegraphics[width=1.1\linewidth,height=3.8cm]{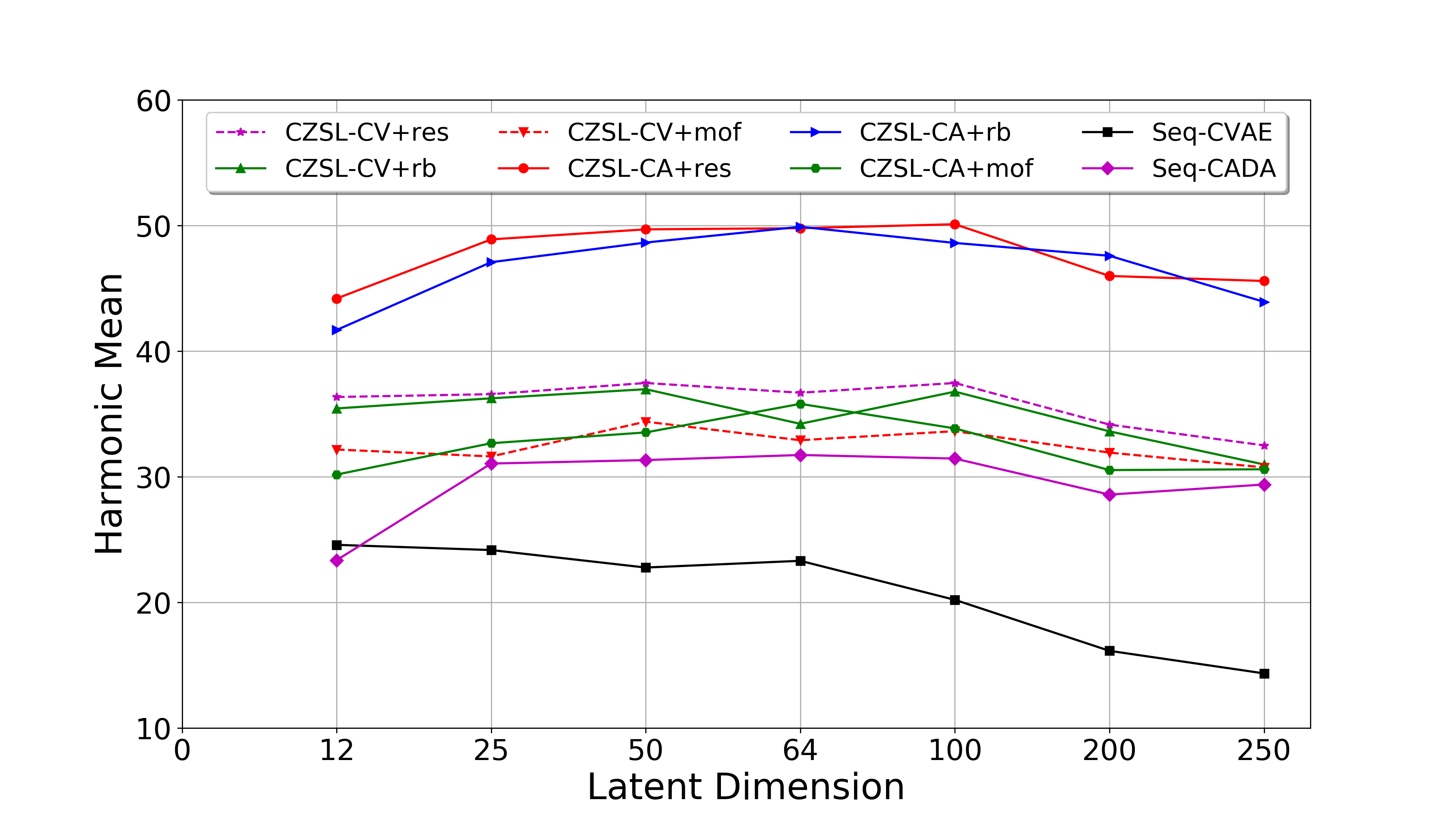}
			\caption{For setting-2}
			\label{fig:abal-lat_s2}
		\end{subfigure}
		\caption{The impact of the latent dimensions of the latent
			features on CUB dataset, which are generated in latent space and utilized for training the linear classifier.}
		\label{fig:abal-lat}
	\end{figure*}

	\subsection{Analyzing Proposed CZSL Method in Detail on CUB Dataset}
	
	\myparagraph{Increasing Episodic Memory Size.} 
	In this analysis, we have experimented with the robustness of the proposed method variants for different samples per class stored in the memory.  It can be analyzed from Figure \ref{fig:abal-mem}, CZSL-CA+res outperforms baselines just by using 3 and 1 sample per class for CZSL setting-1 and setting-2, respectively. In CZSL setting-2, CZSL-CA+res gains more than 10\% when three samples per class are used. Here, It is to be noted that repetitive training on a minimal episodic memory improves the results instead of doing overfitting. As discussed earlier, this performance improvement is owed to data-dependent regularization of $(t+1)^{th}$ task by $t^{th}$ task. Data-dependent regularization can also show an adverse effect on the performance if tasks are not related to each other. It can also be observed in Figure \ref{fig:abal-mem} that the performance improves as the number of samples per class increases. Overall, the reservoir sampling-based method yields the best results among all sampling strategy.       
	
	\myparagraph{Task-wise Performance Analysis.}
	Performance of the proposed CZSL methods mainly depends upon three-factor: (i) the number of examples in memory for any task, and (ii)  task relatedness between samples from the seen and unseen tasks, (iii) task relatedness between the $t^{th}$ and $(t+1)^{th}$ task. For generalized CZSL setting-1 on CUB dataset, it can be seen from Figure \ref{fig:cub-mH_s1}, the performance of CZSL-CA+res is improving as the number of a task is increasing because the number of seen classes increases and the number of unseen classes decreases with task increment. In contrast, generalized CZSL setting-2 exhibits different pattern in Figure \ref{fig:cub-mH_s2} on CUB dataset. Here, the performance of CZSL-CA+res declines initially and then increases as the number of tasks increases. The initial reduction in performance is due to less task relatedness among seen and unseen samples. Therefore, harmonic mean (seen+unseen) and per class unseen accuracy (unseen) for each task follows the same pattern (Unseen class plot for both settings on CUB dataset is available in the supplementary material). As discussed earlier, this setting's last task has an identical train and test data as a standard split. Hence, the proposed methods' performance on the last task with their offline base methods is presented in Table \ref{tab:batch_vs_online}. It can be observed from this table that the proposed CZSL methods have given good results compared to baseline methods; however, there is still much scope of improvement as they lack behind the upper bound.                     
	
	\myparagraph{Increasing Number of Latent Dimensions.}
	In this analysis, we experiment with the robustness of the proposed method against several latent dimensions. We have experimented for different sizes of the latent dimensions and plotted in Figure \ref{fig:abal-lat}. Here, It is observed that a number of latent dimensions increases, the performance improves initially; however; it degrades the performance after latent dimensions=64. When the number of latent dimensions increases, it also increases the degree of freedom. In contrast, the lower dimension provides more compact and discriminative features. We suggest a latent dimensions size: $50-64$.       
		
	\section{Conclusion}
This paper tackles the continual zero-shot learning problem using experience replay and a generative method. The proposed CZSL method performs task-agnostic prediction (single-head) and suitable for class-incremental learning. The performance is evaluated using five ZSL benchmark datasets for two different settings of continual learning with and without class incremental. Our experimental study shows that the experience replay-based CZSL method significantly improves the performance over baseline methods by adding 1 to 3 samples per class in memory. Three types of memory populating strategies have also been studied for two settings of CZSL. Moreover, repetitive training from memory performs data-dependent regularization and improves performance instead of overfitting. The potential future directions are a) understanding the best possible way to populate the memory; b) removing samples from the memory when it is full, and c) develop the task-free CZSL.  
	
	{\small
		\bibliographystyle{ieee_fullname}
		\bibliography{egbib}
	}
	
\end{document}